%% file: root.tex
\documentclass[letterpaper, 10 pt, conference]{ieeeconf}  

\IEEEoverridecommandlockouts                              

\input{preamble}

\title{\LARGE \bf
\acs{MS2MP}: A Min-Sum Message Passing Algorithm for Motion Planning
}

\author{Salman Bari, Volker Gabler and Dirk Wollherr$^{*}$
\thanks{$^{*}$Authors are associated with the Chair of Automatic Control Engineering (LSR), Department of Electrical and Computer Engineering, Technical University of Munich, Germany
        {\tt\small [s.bari|v.gabler|dw]@tum.de}}%
}

\begin{document}

\maketitle
\thispagestyle{empty}
\pagestyle{empty}


\begin{abstract}
\ac{GP} formulation of continuous-time trajectory offers a fast solution to the motion planning problem via probabilistic inference on factor graph. 
However, often the solution converges to in-feasible local minima and the planned trajectory is not collision-free. 
We propose a message passing algorithm that is more sensitive to obstacles with fast convergence time.
We leverage the utility of min-sum message passing algorithm that performs local computations at each node to solve the inference problem on factor graph.
We first introduce the notion of compound factor node to transform the factor graph to a linearly structured graph.
We next develop an algorithm denoted as \ac{MS2MP} that combines numerical optimization with message passing to find collision-free trajectories.
\ac{MS2MP} performs numerical optimization to solve non-linear least square minimization problem at each compound factor node and then exploits the linear structure of factor graph  to compute the \ac{MAP} estimation of complete graph by passing messages among graph nodes.
The decentralized optimization approach of each compound node increases sensitivity towards avoiding obstacles for harder planning problems. 
We evaluate our algorithm by performing extensive experiments for exemplary motion planning tasks for a robot manipulator.
Our evaluation reveals that \ac{MS2MP} improves existing work in convergence time and success rate.
\end{abstract}

\glsresetall

\section{Introduction}\label{sec:introduction}
\vspace{-15cm}
\mbox{\small This~work~has~been~accepted~to~the~IEEE~2021~International~Conference~on~Robotics~and~Automation~(ICRA).}
\vspace{14.2cm}
\vspace{-14.6cm}
\mbox{\small ~~ The~published~version~may~be~found~at~https://doi.org/10.1109/ICRA48506.2021.9561533.}
\vspace{14cm} 

Kinematic motion planning focuses on finding a trajectory in a robot's configuration space from start state to goal state while satisfying multiple performance criteria such as collision avoidance, joint limit constraints and trajectory smoothness.
There are several motion planning approaches that have been proposed so far.
These approaches can be roughly divided into two broad categories: sampling-based algorithms and optimization-based algorithms. 
Sampling-based algorithms \cite{KavrakiSLO96,KuffnerL00,L2006} efficiently find collision free trajectories by probing the configuration space and checking the feasibility of robot's configuration.
However, trajectories produced by sampling-based algorithms are not smooth and therefore require further post processing.

Trajectory optimization algorithms~\cite{RatliffZBS09,KalakrishnanCTPS11,ParkPM12} start with an initial trajectory that may not be collision-free and then minimize an objective function that penalizes collisions and non-smooth configurations.
A drawback of these approaches is that they only tend to find locally optimal solutions and need fine-discritization of trajectory into way points for collision checking in complex environments.
\ac{GP} formulation of continuous-time trajectories~\cite{BarfootTS14} can overcome this challenge. 
\begin{figure}[t]
\centering

\subfloat[]{%
  \includegraphics[width= 0.48 \linewidth]{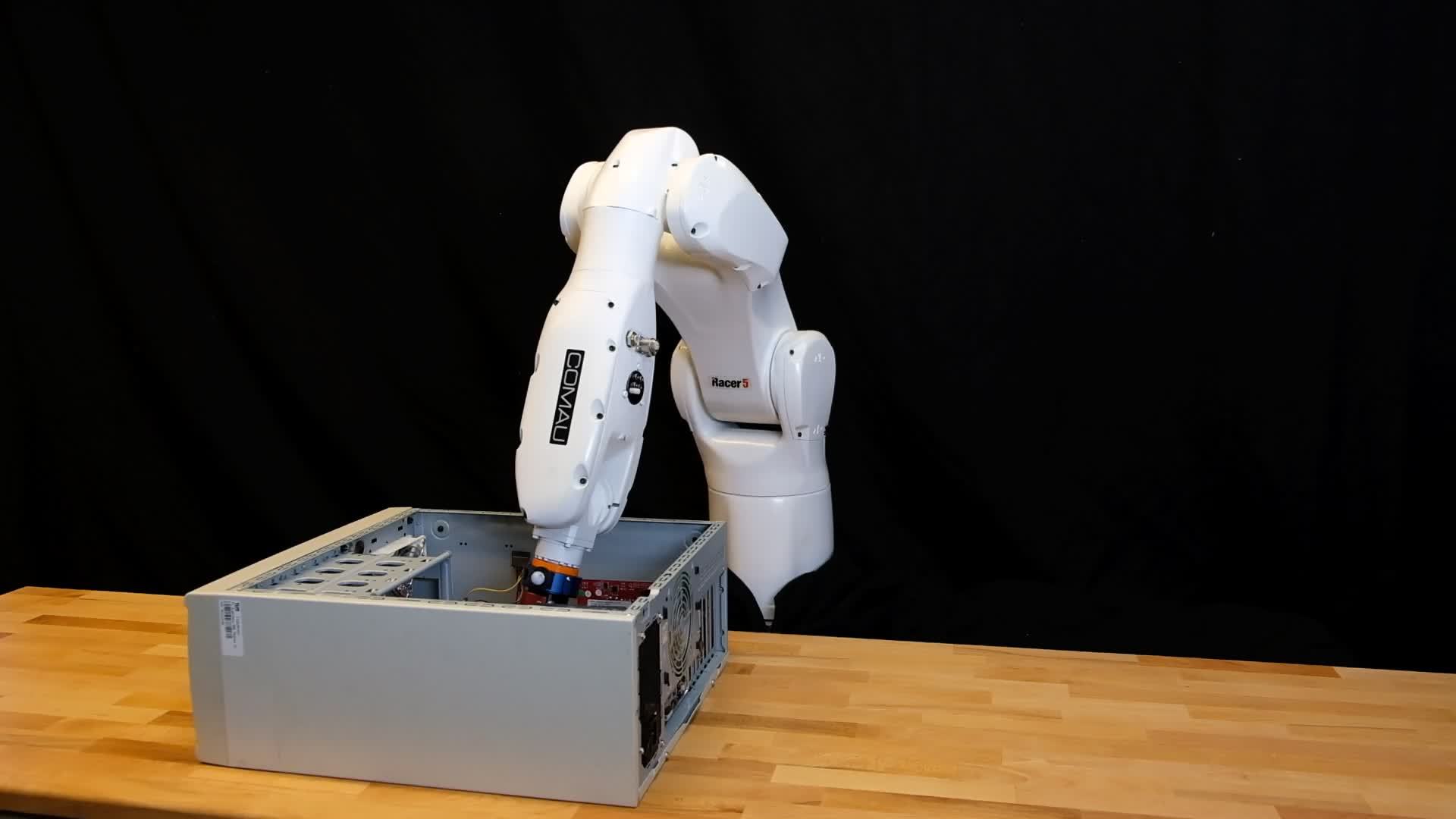}%
  \hskip 1.1ex
  \label{fig0hrdw}%
}
\subfloat[]{%
  \includegraphics[width=0.48 \linewidth]{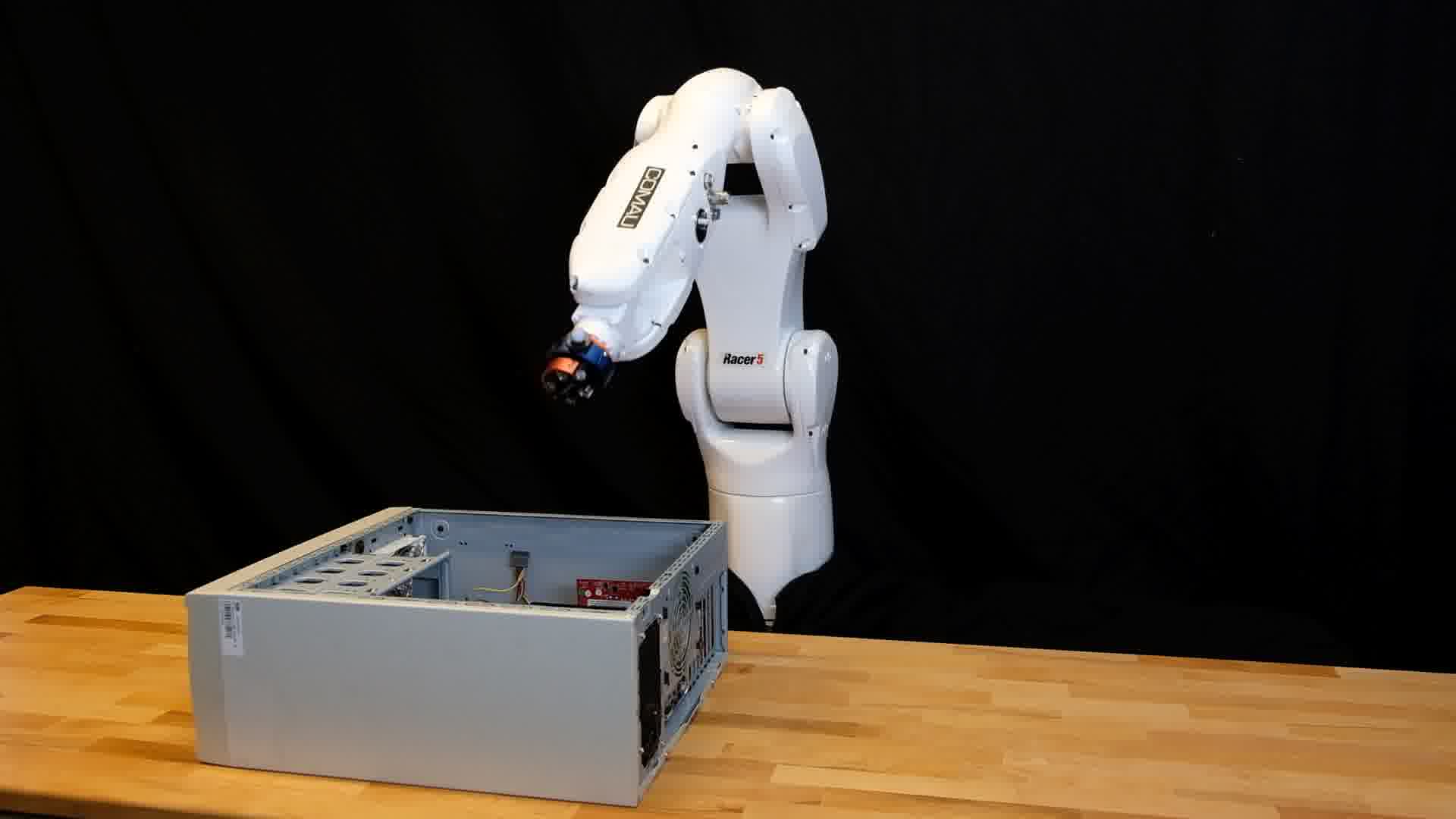}%
   \label{fig4hrdw}
}
\\
\subfloat[]{%
  \includegraphics[width=0.48 \linewidth]{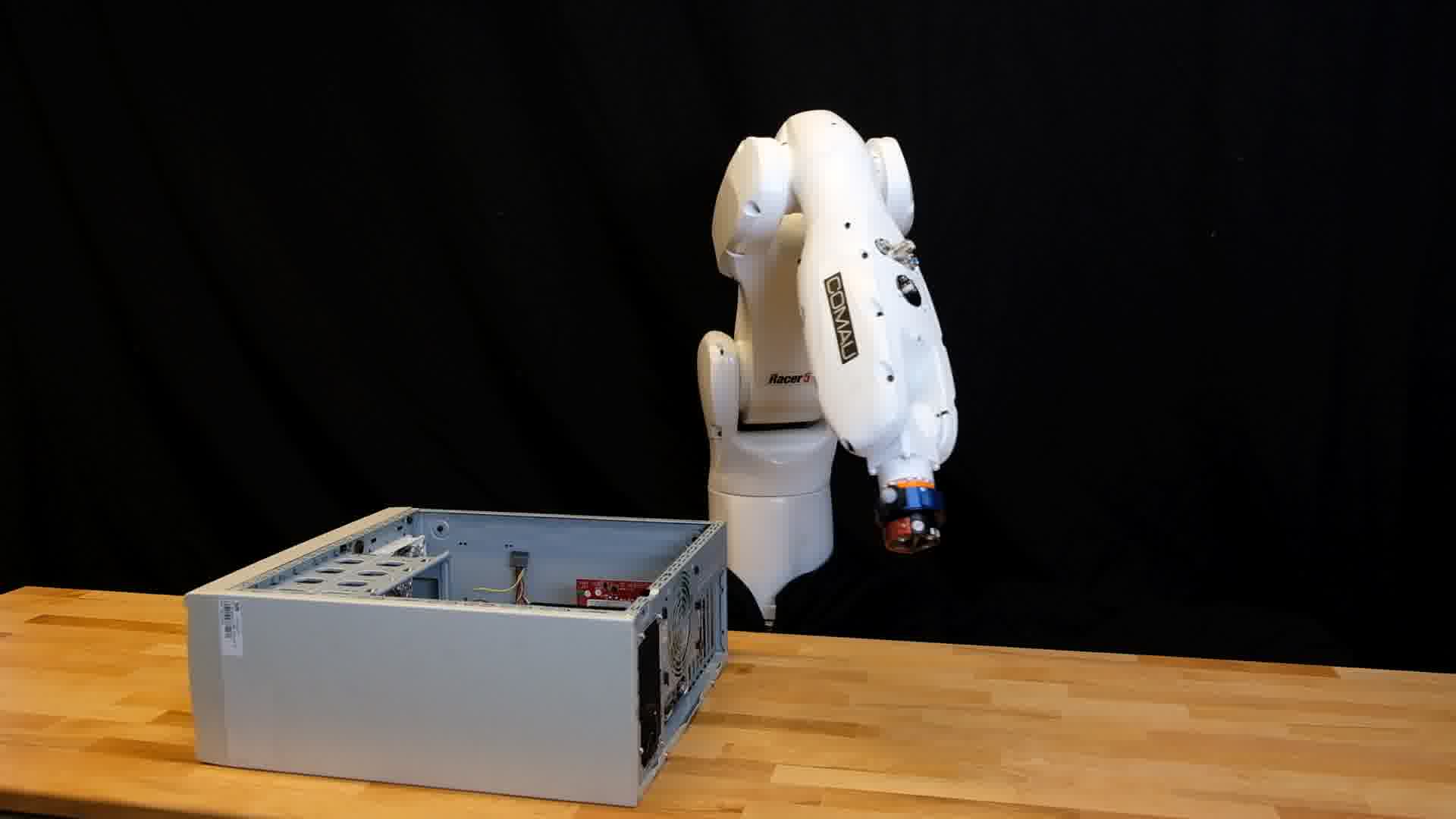}%
   \hskip 1.1ex
  \label{fig7hrdw}%
}
\subfloat[]{%
  \includegraphics[width=0.48 \linewidth]{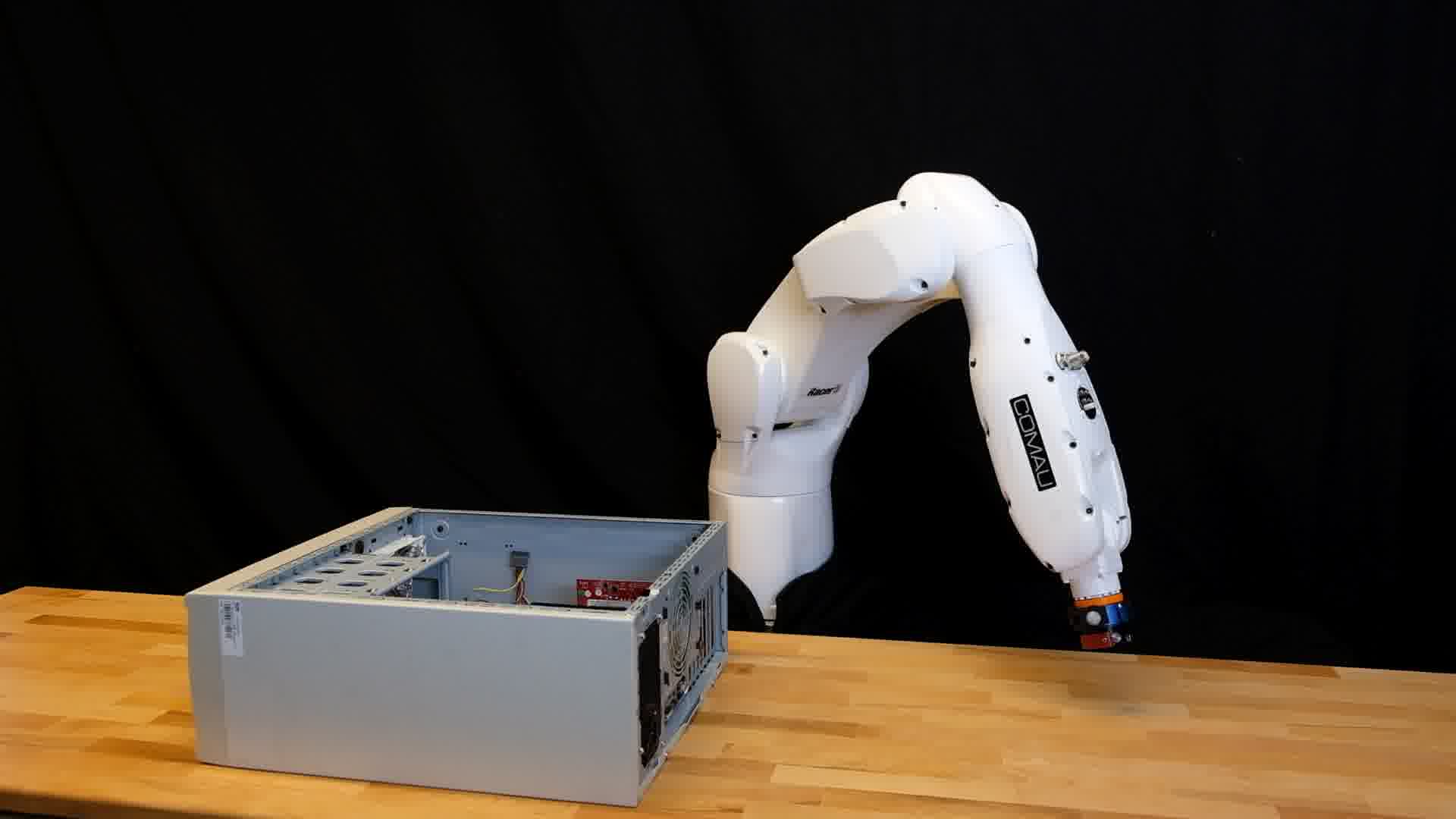}
  \label{fig10hrdw}
}

\caption{Different Robot configurations during MS2MP planning phase for an autonomous disassembly setup: (a) at $N = 0$,  (b) at $N= 4$, (c) at $N = 7$, (d) at $N = 10$.}
\label{fig:experiment:phases}
\end{figure}

Recently, the \ac{GPMP2} framework \cite{MukadamDYDB18} has been proposed that represents trajectory as a \ac{GP} and finds collision-free trajectories with fast, structure-exploiting inference. It fuses all the planning problem objectives which are represented as factors and solves non-linear least square optimization problem via numerical (Gauss-Newton or Levenberg-Marquardt) methods.
Although, \ac{GPMP2} is fast but the batch non-linear least square optimization approach makes it vulnerable to converging to in-feasible local minima.
The approach to combine all the factors of the graph makes it faster than state of the art motion planning algorithms but it comes at the cost of being more prone to getting stuck in local minima.
Graph re-optimization could naively help to get out of the in-feasible local minima but it increases the computation time.

In this work, we explore the min-sum algorithm for finding the \ac{MAP} estimation of a motion planning problem on factor graph.
However, min-sum cannot be efficiently adopted due to the non-linear factors of probabilistic motion models. So, we propose a hybrid algorithm called \ac{MS2MP} that combines numerical optimization methods with min-sum message passing algorithm to solve non-linear factors for \ac{MAP} estimation. 
\ac{MS2MP} combines all the factors attached to the same variable node to form a compound factor node. 
Then, the messages are passed among adjacent factor and variable nodes to find the \ac{MAP} estimation of complete graph that generates collision-free trajectories.


\section{Related Work}\label{sec:relatedWork}
The probabilistic inference view on motion planning is introduced by Toussaint et al.~\cite{Toussaint09,ToussaintG10}. 
The motion planning problem is formulated as a probabilistic model which represents the performance criteria (e.g. avoiding obstacles, reaching a goal) as objectives and use probabilistic inference to compute a posterior distribution over possible trajectories.
Belief propagation is proposed to solve the approximate inference problem of finding feasible trajectories while factor graphs are used to represent the planning problem. 

The \ac{GP} formulation of continuous-time trajectories~\cite{BarfootTS14} made the probabilistic view on motion planning problem more lucrative because it offers sparse parametrization of the trajectory. 
Here, a \ac{GP} is used to represent trajectories as functions that map time to the robot state which provides the benefit of querying the trajectory at arbitrary time steps by exploiting the Markov property of \ac{GP} priors produced from \ac{LTV-SDE}. Recently, B-spline~\cite{ElbanhawiSJ16} and kernel methods~\cite{MarinhoBDBGS16} have been used in a similar manner to represent trajectories with fewer states in motion planning problems.

The formulation of continuous-time trajectory as \ac{GP} paved the way to consider an alternative approach for solving the inference problem on factor graph. 
The \ac{GPMP2} algorithm combines all the factor nodes and solves the inference problem using batch numerical optimization approach. 
Simultaneous Localization and Mapping (SLAM) problems \cite{DellaertK06} have been also solved in the similar manner.
The algorithm often converges to in-feasible local minima due to the holistic approach of solving factor graph and results in trajectories that are not collision-free.
In case of batch optimization, several ideas like random initializations and graph-based initialization~\cite{HuangMLB17} exist to improve results but they do not deal with the inherent approach of combining all factors and optimizing it at once.

Intuitively, the better approach is to evaluate each node of the factor graph separately to obtain improved results.
This is the standard characteristic of message passing algorithms, which in its purest form, allows in-place inference on a factor graph with entirely local processing at graph nodes.
Over the past few years, there has been an increased interest in a message passing based optimization algorithm for graphical models, called min-sum message passing algorithm~\cite{MoallemiR09,Yedidia1}. 
Message passing algorithms have been used to solve NP-hard combinatorial optimization problems~\cite {Gallager62,RichardsonU01}.
The min-sum algorithm is equivalent to the max-product algorithm, and is closely related to belief propagation also known as sum-product algorithm~\cite{YedidiaFW00}.

We build upon the success of min-sum message passing algorithm for general factor graphs and thus give an outline how this method can be adopted for GP-based motion planning problems.
We transform the factor graph by introducing compound factor nodes and then optimize each node locally to generate messages to compute \ac{MAP} estimation of the factor graph.


\section{Problem Formulation}\label{sec:problem}
Suppose a trajectory is represented by $\bm{x}\left ( t \right ) : t\rightarrow \mathbb{R}^{D}$, where D is the dimensionality of the state.
$p\left ( \bm{x} \right )$ is the prior on $\bm{x}$ that actually encourages smooth trajectories and fixes start and goal states, $l\left ( \bm{x };\mathbf{e} \right )$ is the likelihood of states $\bm{x}$ representing collision-free events  $\mathbf{e}$ on the trajectory. 
For example, $e_i$ could be a binary event with $e_i = 0$ if trajectory is collision-free and $e_i = 1$ if trajectory is in collision.
Given the prior and likelihood, the optimal trajectory $\bm{x }^{*}$ is found by the \ac{MAP} estimation, which generates the trajectory that maximizes the conditional posterior $p\left ( \bm{x} | \mathbf{e}\right )$,
\begin{equation}
     \bm{x}^{*} = \underset{\bm{x} }{\arg\max} \;
             \underbrace{p\left ( \bm{x} | \mathbf{e}\right )}_{ p\left ( \bm{x} \right )l\left ( \bm{x };\mathbf{e} \right ) }.
    \label{eq:problem}
\end{equation}
The posterior distribution can be equivalently formulated as \ac{MAP} inference problem on a \emph{factor graph}.
Factor graph formulation for motion planning problem~\eqref{eq:problem} is described in detail in the following section. 

\subsection{Factor Graph Formulation}\label{sec:factorGraphs}
\begin{figure}[t]
    \centering
    \includegraphics[width =\linewidth]{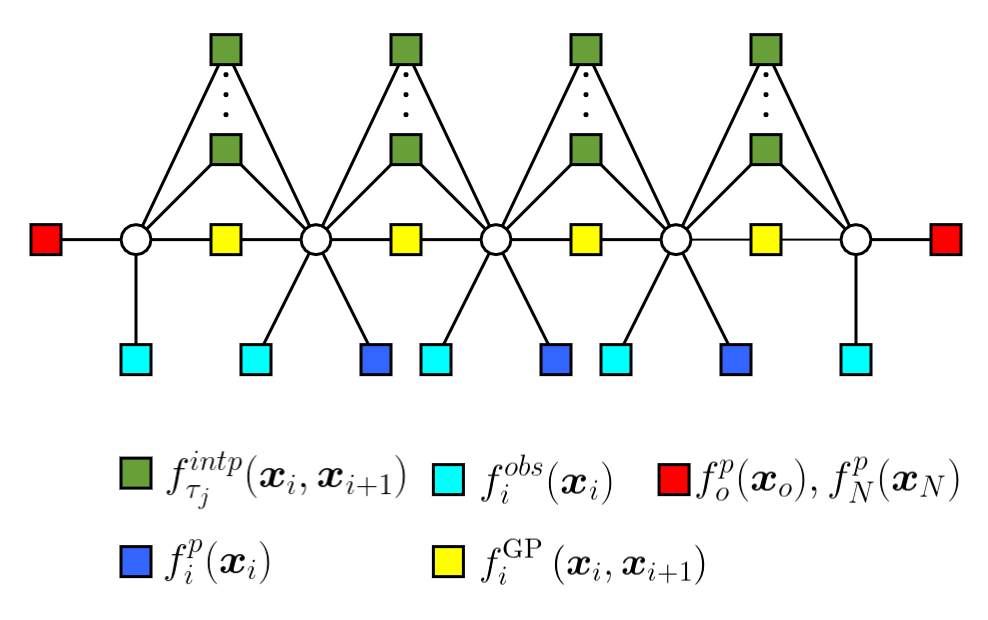}
    \caption{An example trajectory planning problem formulation on a factor graph. White circles show support states and square boxes show factors.}
    \label{fig:factorGraphFormulation}
\end{figure}
The motion planning problem formulation in~\eqref{eq:problem} consists of two components, prior and likelihood. Our objective is to find a trajectory parameterized by $\bm{x}$ that satisfies collision-free events `$\mathbf{e}$'.

The continuous-time trajectory prior $p\left ( \bm{x} \right )$ is represented as functions sampled from \acp{GP} \cite{BarfootTS14}. 
\ac{GP} based trajectory priors offer the benefit of representing the full trajectory $\bm{x}$ by means of $N$ support states $\bm{x }_{i}\in \mathbb{R}^{D}$. 
Given these, a trajectory can be efficiently interpolated between two consecutive support states through Gaussian Process regression (GPR).
Similar to prior work \cite{MukadamDYDB18}, we assume $\bm{x}$ to follow a joint Gaussian distribution
\begin{equation}
    \bm{x }\left ( t \right ) = \left [ \bm{x }_{0} , . . .,\bm{x }_{N} \right ]^{T} \sim 
    \mathcal{N}\left ( \bm{\mu }\left (\bm{t} \right ), \bm{\mathcal{K}}\right ), 
    \label{eq:problem:jointDist}
\end{equation}
for a set of times $\bm{t}=t_0,\dots, t_N$,
where $\bm{\mu }\left ( t \right )$ is a mean vector and $\bm{\mathcal{K}}\left ( t, {t}' \right ) $ is the covariance kernel. 
The kernel $\bm{\mathcal{K}}$ induces smoothness and puts constraints on start and goal states. 
We follow \cite{BarfootTS14} by considering a structured kernel generated from \ac{LTV-SDE}, such that the \ac{GP} prior of $\bm{x}$ results in 
\begin{equation}
p\left ( \bm{x} \right ) \propto \exp \left \{ -\frac{1}{2}\left \| \bm{x -\bm{\mu }} 
\right \|_{\bm{\mathcal{K}}}^{2}
 \right \},
    \label{eq:problem:prior}
\end{equation}
where $\left \| \bm{x -\bm{\mu }},\right \|_{\bm{\mathcal{K}}}^{2}$ is the Mahalanobis distance. 
The likelihood function which specifies the probability of avoiding obstacles is also defined as a distribution in the exponential family. The collision-free likelihood is
\begin{equation}
l\left ( \bm{x };\mathbf{e} \right ) =  \exp \left \{ -\frac{1}{2}\left \| \mathbf{h} \left (\bm{x} \right )
\right \|_{\bm{\Sigma }_{\mathrm{obs}}}^{2}
 \right \},
     \label{eq:problem:likelihood}
\end{equation}
where $\mathbf{h} \left (\bm{x} \right )$ represents the obstacle cost, and $\bm{\Sigma }_{\mathrm{obs}}$ is a hyperparameter matrix.

In the context of~\ac{GP}-based motion planning, we need to model a belief over continuous, multivariate random variables $\bm{x}_i\in \mathbb{R}^{D}$. 
It can be formulated on a factor graph $\mathcal{G} = \left ( \mathcal{X} , \mathcal{F}, \mathcal{E} \right )$ using conditional \acp{PDF} $p\left ( \bm{x}_{i} | \mathbf{e}\right )$ over the variables $\bm{x}_i$,
given the constraints $\mathbf{e}$. 
The bipartite graph $\mathcal{G}$ factorizes the conditional distribution over $\bm{x}$ as
\begin{equation}
 p\left ( \bm{x} | \mathbf{e}\right ) \propto \prod_{m=1}^{M} \;f_{m}\left ( \mathcal{X}_{m} \right ),
     \label{eq:factorGraph:factorization}
\end{equation}
given a subset of $M$ factor nodes $f_{m} \in \mathcal{F}$ on an adjacent subset of variable nodes $\mathcal{X}_{m} \in \mathcal{X}$, 
that are set in relation via the edges $\mathcal{E}$ of the factor graph. 
Fig. \ref{fig:factorGraphFormulation} shows an example factor graph for a motion planning problem with $N=4$ support states that describe a collision-aware motion planning problem.
Thus,  the prior is factorized as
\begin{equation}
 p\left ( \bm{x} \right ) \propto f_{0}^{p}\left ( \bm{x  }_{0} \right ) f_{N}^{p}\left ( \bm{x  }_{N} \right )
 \prod_{i=1}^{N-1}f_{i}^{p}\left ( \bm{x  }_{i}\right )
 f_{i}^{\mathrm{\acs{GP}}}\left ( \bm{x  }_{i}, \bm{x  }_{i+1} \right ),
     \label{eq:factorGraph:prior}
\end{equation}
where $f_{0}^{p}\left ( \bm{x  }_{0} \right )$ and $f_{N}^{p}\left ( \bm{x  }_{N} \right )$ 
put constraints on fixing the start state $\bm{\mu }_{0}$ and goal state $\bm{\mu }_{N}$. 
Constraining the deviation from the prior is included via 
\begin{equation}
\begin{aligned}
f_{i}^{p}\left ( \bm{x  }_{i} \right ) = \exp \left \{ -\frac{1}{2}\left \| \bm{x}_{i} -\bm{\mu }_{i} 
\right \|_{\bm{\mathcal{K}}_{i}}^{2}
 \right \},
     \label{eq:factroGraph:priorConstraint}
\end{aligned}
\end{equation}
in case they are not colliding with obstacles,
while the GP prior of the trajectory is incorporated via
\begin{equation}
\begin{aligned}
f_{i}^{\mathrm{\acs{GP}}}\left(\bm{x  }_{i}, \bm{x  }_{i+1}\right) = 
  \exp \Big\{ & -\frac{1}{2} 
    \left\|\bm{\Phi} (t_{i+1}, t_{i}) \bm{x}_{i}- \bm{x}_{i+1}
    \right.
    \\ & \hspace{25pt}
  \left.\left.+\bm{\mu }_{i, i+1} 
\right \|_{\bm{Q}_{i, i+1}}^{2}
 \right\},
     \label{eq:factorGraph:GPprior}
\end{aligned}
\end{equation}
using state transition matrix $\bm{\Phi}\left ( t_{i+1},t_{i} \right )$ and 
power spectral density matrix ${\bm{Q}_{i, i+1}}$ as introduced in \cite{BarfootTS14}.

The collision-free likelihood $l\left ( \bm{x };\mathbf{e} \right ) $ is factorized with two types of factors, unary obstacle factors $f_{i}^{\mathrm{obs}}$ and interpolated binary obstacle factors $f_{\tau _{j}}^{\mathrm{intp}}$ as
\begin{equation}
l\left ( \bm{x };\mathbf{e} \right ) = \prod_{i=0}^{N} \;\left \{ f_{i}^{\mathrm{obs}}\left ( \bm{x  }_{i}
 \right)\;\prod_{j=1}^{N_{ip}}\;f_{\tau _{j}}^{\mathrm{intp}}\left ( \bm{x  }_{i},\bm{x  }_{i+1}\right)
\right \},
     \label{eq:factorGraph:likelihood}
\end{equation}
where $N_{ip}$ is the number of interpolated states between two consecutive support states $\bm{x }_{i}$, $\bm{x}_{i+1}$ and $\tau _{j}$ is the interpolation time.
The unary obstacle factor at each support state is defined according to~\eqref{eq:problem:likelihood}. 
The interpolated binary obstacle factor is defined as 
\begin{equation}
\begin{aligned}
 & f_{\tau _{j}}^{\mathrm{intp}}\left ( \bm{x  }_{i}, \bm{x  }_{i+1}\right)
\\ & = \exp \left\{ -\frac{1}{2}\left \| \mathbf{h}_{\tau _{j}}^{\mathrm{intp}} \left ( \bm{x  }_{i}, \bm{x  }_{i+1}\right)
\right \|_{\bm{\Sigma }_{\mathrm{obs}}}^{2}
 \right \}.
     \label{eq:factorGraph:likelihoodInter}
\end{aligned}
\end{equation}

The interpolated binary obstacle factor accumulates the collision cost at interpolated states $\tau_{j}$. 
This cost information is utilized to update the associated support states by solving \ac{MAP} inference problem on the factor graph.


\section{MAP Inference via Min-Sum Message passing}\label{sec:MS2MP}
Given $\mathcal{F}$ as a set of Gaussians,~\eqref{eq:factorGraph:factorization} results in a product of Gaussians.
Thus, finding the most probable values for $\bm{x}$ is equivalent to minimizing the negative log of the probability distribution via
\begin{equation}
    \bm{x}^{*}  = \underset{\bm{x} }{\arg\min } \sum_{m=1}^{M} \;f_{m}\left ( \mathcal{X}_{m} \right ).
    \label{eq:logMinSum}
\end{equation}

As the individual factors are nonlinear, it is in-feasible to apply min-sum directly to solve the optimization problem in~\eqref{eq:logMinSum}.
It requires to linearize factors at each step resulting high computation cost.
However, we propose to explicitly alter the factor graph structure by introducing
\emph{compound factor nodes} and then adopt the min-sum algorithm in which a local node objective function is optimized using the Gauss-Newton method.
Our proposed algorithm decreases the tendency of converging to local minima due to distributed approach, while containing the convergence properties, as shown in~\nameref{sec:appendA}. 
In the following, the concept of compound factor nodes is outlined in detail.

\subsection{Factor Graph with Compound Factor Nodes}\label{sec:compoundNode}
\begin{figure*}[t]%
    \centering
    \subfloat[]{{\includegraphics[width=0.32\textwidth]{Images/factorGraphForm.png} }} \subfloat[]{{\includegraphics[width=0.32\textwidth]{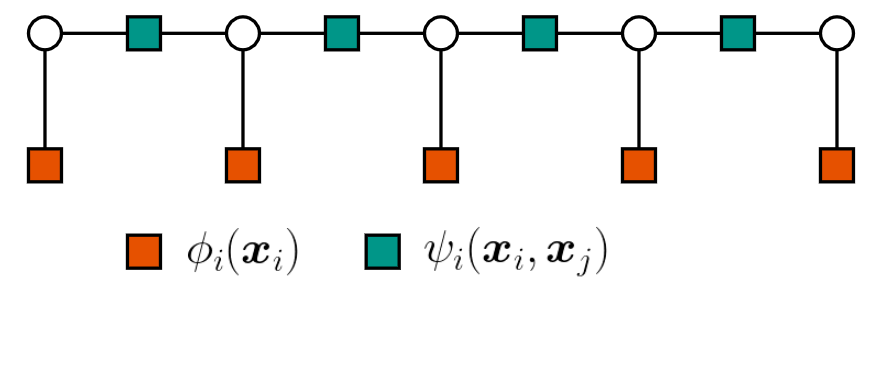} }} \subfloat[]{{\includegraphics[width=0.32\textwidth]{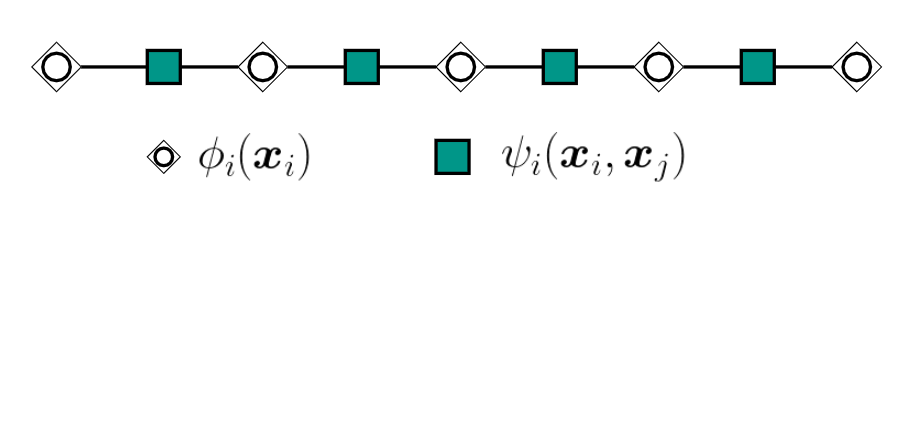} }} 

    \caption{(a) Complete factor graph structure (b) Factors attached to same variable nodes are combined together to form a compound factor nodes. Then, in (c), a linear graph structure is shown. Unary factors attached to variable nodes are not shown explicitly for clarity. Instead, we show that graph consists of self-potentials $\phi_{i }\left ( \bm{x  }_{i} \right )$ and edge-potentials $\psi _{i }\left ( \bm{x  }_{i} , \bm{x  }_{j}\right )$.}

    \label{fig:compoundExample}%
\end{figure*}
The goal of introducing compound factor nodes is to reshape general factor graphs into a linear 
representation with unique edges between nodes by combining individual factors as exemplarily shown in Fig.~\ref{fig:compoundExample} for the graph from Sec.~\ref{sec:factorGraphs}.
\begin{assumption} \label{as:graphType}
A motion planning problem can be fully described by a factor graph $\mathcal{G}$ with $\mathcal{F}$ consisting of two types of factors,
unary factors 
$f_{i} :\mathcal{X} \to \mathbb{R} \;  \cup \; \left \{ \infty  \right \}$ 
and binary factors 
$f_{ij} : \mathcal{X} \times \mathcal{X} \to  \mathbb{R} \;  \cup \; \left \{ \infty  \right \}$ connected to two consecutive variables $\bm{x }_{i}$, $\bm{x }_{j}$ where $\left \{ ij \right \} \in \mathcal{E}$ and $\left |i - j \right | = 1$.
\end{assumption}
\begin{definition}[Compound factor node] \label{def:compoundNode}
Based on assumption~\ref{as:graphType}, a compound unary factor node is defined as
\begin{equation}
    \phi_{i }\left ( \bm{x }_{i} \right ) = \sum_{i\;\in\; \mathcal{X}} \;f_{i}\left ( \bm{x}_{i} \right  ),
    \label{eq:compunds:unary}
\end{equation}
also denoted as self-potential
and binary compound node
\begin{equation}
   \psi _{i }\left ( \bm{x  }_{i} , \bm{x  }_{j}\right ) = \sum_{ij\;\in\; \mathcal{E}} \;f_{ij}\left ( \bm{x  }_{i} , \bm{x  }_{j}\right ),
    \label{eq:compunds:binary}
\end{equation}
as the edge-potential of the current graph.
\end{definition}
Given Definition~\ref{def:compoundNode}, ~\eqref{eq:logMinSum} results in  
\begin{equation}
    \bm{x }^{*}  = \underset{\bm{x } }{\arg\min } \sum_{i} \;\phi_{i }\left ( \bm{x  }_{i} \right ) + \sum_{ij} \;\psi _{i }\left ( \bm{x  }_{i} , \bm{x  }_{j}\right ). 
    \label{eq:compounds:logMinSum}
\end{equation}

The particular linear representation of factor graph, shown on the right hand side of Fig.~\ref{fig:compoundExample} is a pair-wise factor graph. 
Here, min-sum algorithm allows to solve the pair-wise factor graph by iteratively passing messages among nodes.
\subsection{Min-sum Messages Equations}\label{sec:minsum}
\begin{algorithm}[t]

\KwData{trajectory prior $\bm{x}$, 
        factor graph $\mathcal{G}=\left \{ \mathcal{V}, \mathcal{E}\right\}$,
        number of support states $N$
        }
\KwResult{optimized trajectory $\bm{x}^{*}$}
\SetKwFunction{Init}{Init}
\SetKwFunction{LocalUpdate}{LocalUpdate}
\SetKwFunction{BinaryUIpdate}{BinaryUIpdate}
\SetKwProg{UP}{UnitaryUpdate}{}{}
\SetKwProg{BUP}{BinaryUIpdate}{}{}


\Init{$\bm{x}$}\Comment*[r]{Initialize  messages from prior}
\vspace{5pt}
\tcc{Compound Factor Node Formation (Def. \ref{def:compoundNode})}
\For {${i}\in\mathcal{X}$}{
     $\phi_{i}\left ( \bm{x}_{i} \right  )\;\;$ $\leftarrow$ $\;\;{\sum}_{i\in\mathcal{X}} \;f_{i} \left( \bm{x}_{i} \right ) \;$
}
\For {${ij}\in\mathcal{E}$}{
     $\psi _{i }\left ( \bm{x  }_{i} , \bm{x  }_{j}\right ) = \sum_{ij\;\in\; \mathcal{E}} \;f_{ij}\left ( \bm{x  }_{i} , \bm{x  }_{j}\right)\;$
}

\vspace{5pt}
\tcc{Min-sum messages calculation}
\For{$t=0,1,\dots N-1$}{
    \tcc{get variable to factor messages}
        $m_{x\to f}^{t}\left (\bm{x}_{i} \right) \&= 
  \phi_{i }\left (\bm{x }_{i} \right) + 
  m_{\left (\mathcal{F}_x/f \right) \to x}^{t-1}\left ( \bm{x}_{i} \right ),$
    
    \tcc{get factor to variable messages}
    $m_{f\rightarrow x}^{t}\left ( \bm{x}_{i} \right ),\forall \psi _{i }\left ( \bm{x  }_{i} , \bm{x  }_{j}\right ) \in\mathcal{F} $  \Comment*[r]{see~\eqref{eq:message:update:factor}}
    \tcc{update (belief) state}
    $b_{i}^{t}\left (\bm{x}_{i} \right ) \gets  \phi_{i }\left ( \bm{x  }_{i} \right ) + \sum_{f \;\in\; \mathcal{F}_x} 
         m_{f\mapsto x}^{t}\left ( \bm{x}_{i} \right )$\;
    $ \bm{x}_{i} \gets \underset{\bm{x}_{i}}{\arg\min}\;\; b_{i}^{t}\left ( \bm{x}_{i} \right ) \;$
}
return $\bm{x}^{*}\;\;$

\caption{MS2MP: Min-Sum Message Passing Algorithm for Motion Planning}
\label{alg:MSMP@:main}
\end{algorithm}
In the min-sum algorithm, each node solves a local optimization problem and traverses messages to the adjacent nodes.
Namely, we differentiate between messages passed from variables to factor nodes, denoted as 
$m_{x\to f}^{t}$,
and messages passed form factor nodes to variable nodes, denoted as 
$m_{f\to x}^{t}$. 
Denoting the adjacent nodes of a factor node as $\mathcal{X}_f$ and the adjacent nodes of a variable node as $\mathcal{F}_x$, 
the messages are iteratively updated according to 
\begin{align}
m_{x\to f}^{t}\left (\bm{x}_{i} \right) &= 
  \phi_{i }\left (\bm{x }_{i} \right) + 
  m_{\left (\mathcal{F}_x/f \right) \to x}^{t-1}\left ( \bm{x}_{i} \right ),
\label{eq:message:update:variable}
\\
m_{f\to x}^{t}\left ( \bm{x}_{i} \right ) &= \underset{\bm{x}_{j}}{\min }\left [ \psi _{i }\left ( \bm{x }_{i} , \bm{x  }_{j}\right ) +
m_{x \to f}^{t-1}\left ( \bm{x}_{j} \right ) \right ],
\label{eq:message:update:factor}
\end{align}
where $\left (\mathcal{F}_x/f \right)$ notates the set-theoretic difference, i.e. the set of factor nodes $\mathcal{F}_x$ except the factor $f$.
From Algorithm~\ref{alg:MSMP@:main}, at time-step $t=0$, all the messages from variable nodes to factor nodes are initialized according to the prior in~\eqref{eq:message:update:variable}.
In the next step, the min-marginal of $\psi _{i }\left ( \bm{x }_{i} , \bm{x  }_{j}\right )$ is calculated in~\eqref{eq:message:update:factor}.
 Given the messages from adjacent factor nodes, the belief of a variable node is approximated via 
\begin{equation}
b_{i}^{t}\left (\bm{x}_{i} \right ) = 
\phi_{i }\left ( \bm{x  }_{i} \right ) + \sum_{f \;\in\; \mathcal{F}_x} 
\; m_{f\to x}^{t}\left ( \bm{x}_{i} \right ).
    \label{eq:message:update:belief}
\end{equation}
The minima of decision variable $\bm{x}_{i}$ at $t^{th}$ iteration is
\begin{equation}
\bm{x}_{i}^{t} \propto \underset{\bm{x}_{i}}{\arg\min}\;\; b_{i}^{t}\left ( \bm{x}_{i} \right ).
    \label{eq:message:min:belief}
\end{equation}

At each iteration $t$, each node optimizes a local objective function by merging incoming messages into the node.

We use numerical optimization similar to \cite{MukadamDYDB18,DellaertK06} to solve the local non-linear objective function.
Since, each local objective function is composed of multiple factors,~\eqref{eq:message:update:factor} and ~\eqref{eq:message:min:belief} take the form of non-linear least-square problem as
\begin{equation}
\begin{aligned}
m_{f\to x}^{t}\left ( \bm{x}_{i} \right ) =& \underset{\bm{x}_{j}}{\min }
 \Bigg[ \sum_{ij\;\in\; \mathcal{E}}\;
    \left\{ \frac 12
       \left \| \mathbf{h}_{\tau _{j}}^{\mathrm{intp}} 
         \left( \bm{x }_{i}, \bm{x  }_{j}
         \right)
       \right \|_{\bm{\Sigma }_{\mathrm{obs}}}^{2}
    \right \}
  \\& + \sum_{ij\;\in\; \mathcal{E}}\;
  \left\{ \frac 12
    \left \| f_{ij}^{\mathrm{\acs{GP}}} 
      \left ( \bm{x  }_{i}, \bm{x  }_{j}
      \right)
    \right \|_{\bm{Q}_{i, j}}^{2}
  \right \} 
  \\& + m_{x \to f}^{t-1}
   \left ( \bm{x}_{j} \right ) 
   \Bigg],
\label{eq:leastSquare:factor}
\end{aligned}
\end{equation}

\begin{equation}
\begin{aligned}
\bm{x}_{i}^{*} =& \underset{\bm{x}_{i}}{\arg\min}
\Bigg[ \sum_{i\;\in\; \mathcal{X}}\;
    \left \{ \frac 12
        \left \| \bm{x}_{i} -\bm{\mu }_{i} 
        \right \|_{\bm{\mathcal{K}}_{h}}^{2}
    \right \}
    \\ & + \sum_{i\;\in\; \mathcal{X}}\;
    \left\{ \frac{1}{2}
        \left \| \mathbf{h}_{i} 
            \left (\bm{x}_{i} 
            \right )
        \right \|_{\bm{\Sigma }_{\mathrm{obs}}}^{2}
     \right \}
    \\ & + \sum_{f \;\in\; \mathcal{F}_x}\; m_{f\to x}^{t}
    \left ( \bm{x}_{i} \right )
\Bigg].
\label{eq:leastSquare:belief}
\end{aligned}
\end{equation}

We use the Gauss-Newton algorithm to solve the optimization problems in~\eqref{eq:leastSquare:factor} and~\eqref{eq:leastSquare:belief}. 
Note that our proposed algorithm is different in two aspects from \ac{GPMP2} \cite{MukadamDYDB18}: first, the batch optimization of complete graph is replaced by an iterative local optimization of each node, secondly, message passing is performed to achieve the optimization of the complete factor graph.
This behaviour is in fact helpful in avoiding collisions for planning problems when robot has to find its way out of obstacle-rich environment as we empirically show in the experimental validation in the following section.
\begin{figure}[t]
    \centering
    \includegraphics[width = \linewidth]{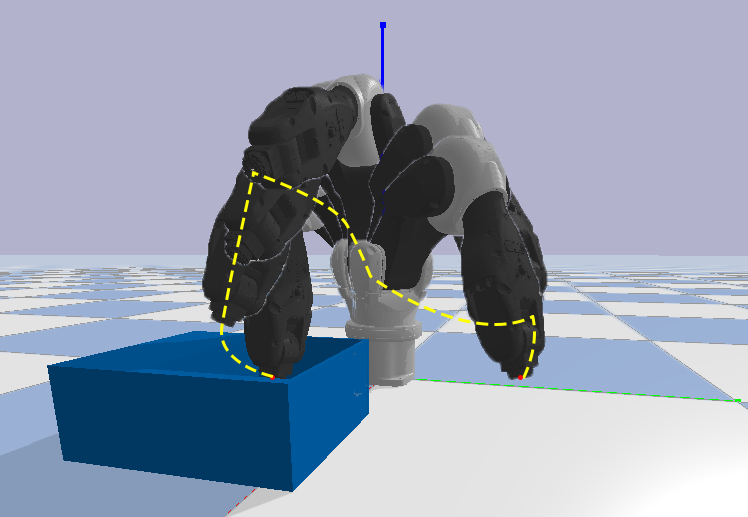}
    \caption{Planned trajectory using \ac{MS2MP} for a 6 \ac{DOF} COMAU Racer 5 
robot.}
    \label{fig:experiment:sim}
\end{figure}

\section{Evaluation} \label{sec:eval}
We evaluated our algorithm on an exemplary motion planning task in a lab environment.
Precisely, a  6 \ac{DOF} COMAU Racer 5 robot is tasked to find collision-free trajectory from the inside of the body of a PC-tower as part of an autonomous disassembly process of Electrical and Electronic equipment. The applicability of the proposed algorithm is also evaluated on the robot platform visualized in Fig.~\ref{fig:experiment:phases}.
We ran benchmark evaluations against~\ac{GPMP2} in a simulated scenario.
\footnote{For a detailed comparison of~\ac{GPMP2} against recent state-of-the-art motion planners, we refer the reader to~\cite{MukadamDYDB18},
where an extensive benchmark comparison has been outlined already.}
The simulation scenario is shown in Fig.~\ref{fig:experiment:sim}, where the PC-tower is approximated by an occupancy mesh. 
In this section, we discuss the implementation and evaluation details of our algorithm.

\subsection{GP Prior} \label{sec:eval:prior}
We use a constant-velocity prior similar to  \cite{MukadamDYDB18} with the Markovian state consisting of configuration position and velocity.
The trajectory is generated from \ac{LTV-SDE}~\cite{BarfootTS14}.
The prior factors for the experiment are presented in~\eqref{eq:factorGraph:prior}.

\subsection{Collision-free likelihood}\label{sec:eval:likelihood}
Similar to recent state-of-the art motion planning algorithms, e.g. \cite{RatliffZBS09,MukadamDYDB18}, the robot collision body is represented by a set of spheres.
We formulate the collision-free unary and binary factors following~\eqref{eq:factorGraph:likelihood}. 
The obstacle cost function is then obtained by computing the hinge loss 
\begin{equation}
\mathbf{c}\left ( d \right )= \left\{\begin{matrix}
-d+\epsilon  & \text{if} \;\; d\leq \epsilon \\ 
 0&  \text{if} \;\; d>  \epsilon
\end{matrix}\right.,
    \label{eq:hinge}
\end{equation}
for the spheres. 
In~\eqref{eq:hinge}
$d \left (x \right)$ represents the signed distance from the sphere to the closest obstacle surface in the workspace, and $\epsilon$ is a \emph{safety distance}.
$\epsilon$  increases the sensitivity of the obstacle constraint before collision can occur. Non-zero obstacle-cost addition enables the robot to not reach too close to the obstacles.
For likelihood in~\eqref{eq:problem:likelihood}, the parameter $\Sigma_{\mathrm{obs}}$ is defined as
\begin{equation}
\bm{\Sigma} _{\mathrm{obs}} = \sigma_{\mathrm{obs}}^{2}\mathbf{I}
    \label{eq:papar:sigma}
\end{equation}
where the parameter $\sigma_{\mathrm{obs}}$ is used to add weight to the \emph{obstacle cost}.

\subsection{Experiment Setting}\label{sec:eval:experiment}
Prior trajectory is initialized by a constant velocity straight line from start state to goal state. 
We initialized the trajectory with $N=11$ support states and 4 additional interpolated states between two consecutive support states.
In our benchmarks we set  $\epsilon = 0.2$ for COMAU Racer arm. 
The term $\sigma_{\mathrm{obs}}$ puts weight on observing the obstacles and it is set according to the planning problem.
we set $\sigma_{\mathrm{obs}}= 0.001$ for our manipulator planning tasks.

We evaluated the proposed algorithm for 24 unique planning problems for different start configurations inside the PC tower. 
To make the planning problem much harder the start configurations have been kept very close to the body of PC tower.
\ac{MS2MP} extends the Matlab toolbox from~\ac{GPMP2} and has thus been benchmarked using identical framework.
The benchmarks have been run on a 3.90GHz Intel Core i3-7100 CPU. 

\subsection{Discussion}\label{sec:eval:analysis}
\begin{table}[b]
\centering
\caption{Results of 24 planning problems for 6-DOF COMAU Racer robot}
\label{eval:results}
\begin{tabular}{@{}lllll@{}}
\toprule
                     & \multicolumn{1}{r}{Success (\%) } & Average time (\si{\second}) & Max. time (\si{\second})      \\ \midrule
\ac{MS2MP}           & 83.3                    & \textbf{0.1028}               & \textbf{0.1751}              \\
\ac{MS2MP}-no-comp   & \textbf{91.6}           & 0.5625                        & 0.8213             \\
\ac{GPMP2}           & 70.8                    & 0.3772                        & 0.3962             \\
\ac{GPMP2}-no-intp   & 66.7                    & 0.2724                        & 0.2873             \\  \bottomrule
\end{tabular}
\end{table}
Table \ref{eval:results} summarizes results for 24 planning problems where a robot has to find its way out of an obstacle-rich environment.
\ac{MS2MP} is more successful in finding collision-free trajectories compared to \ac{GPMP2} with only approximately a third of the run-time.
\ac{GPMP2} leads to an early termination of the optimization algorithm due to an increase in absolute error. 
In this case, a re-optimization of the graph is required in order to naively overcome in-feasible local minima. 
It results in increased run-time for finding collision-free trajectories. However, \ac{GPMP2} will still be faster with less success rate if we do not consider the re-optimization step.

We observe that local optimization step at each compound node increases the sensitivity of obstacle avoidance resulting in better success rate in case of \ac{MS2MP}. 
A drawback of this approach is that the increased sensitivity towards obstacles affects the trajectory smoothness. 
In order to overcome this drawback, an additional unary prior factor is introduced in~\eqref{eq:factorGraph:prior} that induces smoothness in case of \ac{MS2MP}.
\ac{MS2MP}-no-comp has the highest success rate because each non-linear factor is linearized at every message passing step. 
It results in high computation cost, almost double the run-time as compared to \ac{GPMP2} and the output trajectory is not very smooth.
We also benchmarked our algorithm against \ac{GPMP2} without interpolation (\ac{GPMP2}-no-intp) with the same number of support states and no interpolated states. 
Although, its faster than \ac{GPMP2}, the success rate is lowest among all the evaluated algorithms.
We refer the reader to the accompanying video that shows a trajectory planned by \ac{MS2MP}.


\section{Conclusions}\label{s6}

Formulation of motion planning problems on factor graphs has set the stage for applying different approaches to generate feasible and smooth trajectories. 
We build upon the idea of \ac{GPMP2} using factor graphs in order to represent a motion planning inference problem. 
However, a drawback of existing work is that by fusing all factors in the graph and solving a global nonlinear least square problem tends to converge to in-feasible local minima.
This tendency increases with the complexity of motion planning problem.
We believe that this problem can be avoided using message passing  techniques.

Message passing is a strong algorithmic framework for solving \ac{MAP} estimation problem on factor graphs by performing local computations at each node.
We proposes an algorithm called \ac{MS2MP} for finding collision-free trajectories. It performs local computations at individual nodes, thus decreases the chances of converging to in-feasible local minima.
A major benefit of \acs{MS2MP} is that it can be extended to an incremental method that allows planning in dynamic environments, where each node is optimized online. 
An added benefit of message passing approach is the possibility of performing parallel computation acorss the graph nodes that can further reduce planning time significantly. 
For future work, we are considering to investigate the incremental planning and parallel computation of the proposed algorithm.

\addtolength{\textheight}{-10.5cm}   

\section*{Appendix A} \label{sec:appendA}
\subsection{Convergence Analysis}
The min-sum algorithm converges to the global optimum for tree-structured graphs.
However, in our proposed algorithm min-sum message passing differs in two aspects. 
First, we introduce compound factor nodes by merging factors connected to same variable node(s).
Secondly, we have non-linear factors. 
The crucial point in solving the inference problem for non-linear factors is that at each iteration $t$, $\phi_{i }$ and $\psi _{i}$ form local objective functions by combining incoming messages into the node. 
The \ac{MAP} estimation using the min-sum algorithm often fails to converge if the single node local optimization fails to find a unique solution. 
Its not possible to construct $\bm{x}^{*}$ directly if the local node objective function has more than one optimal solution~\cite{WainwrightJW04}.
For this reason, we assume that the local node objective function has a unique minimum. 
\begin{assumption} \label{app:assump:local}
For each node with self-potential $\phi_{i }\left ( \bm{x  }_{i} \right )$ and edge-potential $\psi _{i }\left ( \bm{x}_{i} , \bm{x  }_{j}\right )$, 
where $i\in\mathcal{X}$ and $ij\in\mathcal{E}$, 
the solution produced by numerical optimization of local node objective function always converges to a unique optimum point.
\end{assumption}
Based on Assumption~\ref{app:assump:local}, the obtained beliefs are the min-marginals of the function $\psi _{i }\left ( \bm{x }_{i} , \bm{x  }_{j}\right )$. Similar to max-marginal local optimality condition \cite{WainwrightJW04}, we can also define local optimality for the compound factor nodes in the graph.

\begin{definition}[Local optimality - compound factor nodes] \label{app:def:localOpt}
Solution obtained from a local node objective function is locally optimal (min-consistent) according to \cite{RuozziarXiv} if for all nodes  $\phi_{i }\left ( \bm{x }_{i} \right )$,
$ \psi _{i }\left ( \bm{x  }_{i} , \bm{x}_{j}\right )$
\begin{equation}
\underset{\bm{x}_{j}}{\min }\left [ \psi _{i }\left ( \bm{x }_{i} , \bm{x  }_{j}\right ) + \sum_{x \;\in\; \mathcal{X}_f}\;
m_{x \to f}^{t}\left ( \bm{x}_{i} \right ) \right ] =b^{t}_{i}\left ( \bm{x}_{i} \right ).
    \label{40b}
\end{equation}
\end{definition}

\begin{remark}\label{app:remark:elim}
For a factor graph $\mathcal{G}$, if the local node $\phi_{i}$ optimization converges to unique solution $\bm{x}^{*}_{i}$, then the node $i$ is eliminated from the overall graph with the remaining graph $\mathcal{G}/ \left |{i} \right|$ \cite[proposition 1]{WainwrightJW04}. 
Proceeding in the same manner will result in the the optimized $\bm{x}^{*}$.
\end{remark}
\begin{proposition}
For a linear-structured factor graph $G$ with compound factor nodes, if there exists a unique $\bm{x}_{i}^{*}$ which is the optimum point for node 
$\phi_{i}\in\mathcal{X}$ 
then for the global objective function~\eqref{eq:compounds:logMinSum}, the min-sum message passing algorithm converges to a $\bm{x}^{*}$ together which optimizes entire graph.
\end{proposition}

\begin{proof}
In order for the full graph $\mathcal{G}$ converging to a solution, the set of open nodes in the graph needs to hold $\mathcal{G}_\mathrm{check}=\emptyset$, where $\mathcal{G}_\mathrm{check}$ is initialized by all nodes in the graph.
Thus, given the linear-structured factor graph $\mathcal{G}$, where each node possesses self and egde-potential functions, 
the iterative optimization 
of the min-sum algorithm is run in a monotonically increasing or decreasing manner.
Based on Remark~\ref{app:remark:elim}, 
individual nodes are removed from $\mathcal{G}_\mathrm{check}$
in an ordered sequence, thus
eventually obtaining $\mathcal{G}_\mathrm{check}=\emptyset$ for 
for $t\rightarrow \infty$.
Referring to Definition~\ref{app:def:localOpt}, the obtained graph directly allows to obtain the converged trajectory.
\end{proof}


\section*{ACKNOWLEDGMENT}
The research leading to these results has received funding from the Horizon 2020 research and innovation programme under grant agreement 
820742 of the project ''HR-Recycler - Hybrid Human-Robot RECYcling plant for electriCal and eLEctRonic equipment''.


\bibliographystyle{IEEEtran}

\bibliography{IEEEabrv,mybibfile}

\end{document}

%% file: preamble.tex
\usepackage{float}
\usepackage{graphicx}
\usepackage{subfig}
\usepackage[ruled,vlined]{algorithm2e}
\usepackage{algpseudocode}
\usepackage[stable]{footmisc}
\usepackage[english]{babel}
\usepackage{caption}
\usepackage{hyperref}
\usepackage{booktabs}
\usepackage{amsfonts,amssymb,amsmath,amsgen,amsopn,amsbsy,theorem,graphicx,epsfig}
\usepackage{cite}
\usepackage{amsmath,amssymb,amsfonts,bm}
\usepackage{comment}
\usepackage{xcolor}
\usepackage{siunitx}

\definecolor{TUMblue}{RGB}{0, 101, 189}

\usepackage[acronym, toc=false, shortcuts, nomain, nonumberlist]{glossaries}
\newacronym{GP}{GP}{Gaussian Process}
\newacronym{LTV}{LTV}{linear time-variant}
\newacronym{GPMP2}{GPMP2}{Gaussian  Process  Motion  Planning}
\newacronym{MAP}{MAP}{maximum a posteriori}
\newacronym{LTV-SDE}{LTV-SDE}{linear  time  variant,  stochastic differential equation}
\newacronym{PDF}{PDF}{ probability density function}
\newacronym{MS2MP}{MS2MP}{Min-sum  Message Passing  algorithm  for  Motion  Planning}
\newacronym{DOF}{DOF}{degree-of-freedom}

\makeglossaries

\SetEndCharOfAlgoLine{}
\SetKwComment{Comment}{}{}

\SetCommentSty{mycommfont}

\newtheorem{assumption}{Assumption}
\newtheorem{definition}{Definition}
\newtheorem{remark}{Remark}
\newtheorem{proposition}{Proposition}